\documentclass[journal]{IEEEtran}
\usepackage{cite}

\ifCLASSINFOpdf
  \usepackage[pdftex]{graphicx}
  \usepackage{epstopdf}
  \usepackage{caption}
  \usepackage{subcaption}
\else
  \usepackage[dvips]{graphicx}
  \usepackage{caption}
  \usepackage{subcaption}
\fi
\usepackage{amsmath}
\usepackage{amssymb}
\usepackage{array}
\usepackage{booktabs}

\usepackage{multirow}
\begin{document}


\title{Multi-source Transfer Learning with Convolutional Neural Networks for Lung Pattern Analysis}


\author{Stergios~Christodoulidis,~\IEEEmembership{Member,~IEEE,}
        Marios~Anthimopoulos,~\IEEEmembership{Member,~IEEE,}
        Lukas~Ebner,
        Andreas~Christe,
        and~Stavroula~Mougiakakou*,~\IEEEmembership{Member,~IEEE}

%
%

\thanks{This research was carried out within the framework of the IntACT research project, supported by Bern University Hospital,``Inselspital'' and the Swiss National Science Foundation (SNSF) under Grant 156511. S. Christodoulidis and M. Anthimopoulos contributed equally to this work. Asterisk indicates
corresponding author.}
\thanks{S. Christodoulidis is with the ARTORG Center for Biomedical Engineering
	Research, University of Bern, 3008 Bern, Switzerland (e-mail:
	stergios.christodoulidis@artorg.unibe.ch).}
\thanks{M. Anthimopoulos is with the ARTORG Center for Biomedical Engineering
Research, University of Bern, 3008 Bern, Switzerland, and with the Department of
Diagnostic, Interventional and Pediatric Radiology, Bern University Hospital
``Inselspital'', 3010 Bern, Switzerland, and also with the Department of Emergency
Medicine, Bern University Hospital ``Inselspital'', 3010 Bern, Switzerland
(e-mail: marios.anthimopoulos@artorg.unibe.ch).}
\thanks{L. Ebner and A. Christe are with the Department of Diagnostic,
Interventional and Pediatric Radiology, Bern University Hospital ``Inselspital'',
3010 Bern, Switzerland (e-mails: lukas.ebner@insel.ch; andreas.christe@insel.ch).}
\thanks{S. Mougiakakou* is with the Department of Diagnostic, Interventional
and Pediatric Radiology, Bern University Hospital ``Inselspital'', 3010
Bern, Switzerland, and also with the ARTORG Center for Biomedical Engineering Research, University of Bern, 3008 Bern Switzerland (e-mail:
stavroula.mougiakakou@artorg.unibe.ch).}}

%

\maketitle

\begin{abstract}

Early diagnosis of interstitial lung diseases is crucial for their treatment, but even experienced physicians find it difficult, as their clinical manifestations are similar. In order to assist with the diagnosis, computer-aided diagnosis (CAD) systems have been developed. These commonly rely on a fixed scale classifier that scans CT images, recognizes textural lung patterns and generates a map of pathologies. In a previous study, we proposed a method for classifying lung tissue patterns using a deep convolutional neural network (CNN), with an architecture designed for the specific problem. In this study, we present an improved method for training the proposed network by transferring knowledge from the similar domain of general texture classification. Six publicly available texture databases are used to pretrain networks with the proposed architecture, which are then fine-tuned on the lung tissue data. The resulting CNNs are combined in an ensemble and their fused knowledge is compressed back to a network with the original architecture. The proposed approach resulted in an absolute increase of about 2\% in the performance of the proposed CNN. The results demonstrate the potential of transfer learning in the field of medical image analysis, indicate the textural nature of the problem and show that the method used for training a network can be as important as designing its architecture.

\end{abstract}




%

\section{Introduction}

\IEEEPARstart{I}{NTERSTITIAL} lung diseases (ILDs) include more than 200 chronic lung disorders characterized by inflammation of the lung tissue, which often leads to pulmonary fibrosis. Fibrosis progressively reduces the ability of the air sacs to capture and carry oxygen into the bloodstream and eventually causes permanent loss of the ability to breathe. Early diagnosis of such diseases is crucial for making treatment decisions, while misdiagnosis may lead to life-threatening complications~\cite{society1999diagnosis}. Although ILDs are histologically heterogeneous, they mostly have similar clinical manifestations, so that differential diagnosis is challenging even for experienced physicians. High resolution computed tomography (HRCT) is considered the most appropriate protocol for screening ILDs, due to the specific radiation attenuation properties of the lung tissue. The CT scans are interpreted by assessing the extent and distribution of the existing ILD pathologies in the lung. However, the inherent difficulty of the problem and the large quantity of radiological data that radiologists have to scrutinize result in low diagnostic accuracy and high inter- and intra-observer variability, which may be as great as 50\%~\cite{Sluimer2006}. This ambiguity in the radiological assessment often leads to additional histological biopsies which increase both the risk and cost for patients. In order to assist the radiologist with the diagnosis and to avoid biopsies, a lot of research has been done towards computer-aided diagnosis (CAD) systems. The basic module of such systems is often a fixed scale texture classification scheme that detects the various ILD patterns in the CT scan and outputs a map of pathologies, which is later used to reach a final diagnosis. To this end, a great variety of image descriptors and classifiers have been proposed for recognizing lung patterns.

Deep learning techniques and especially convolutional neural networks (CNNs) have attracted much attention after the impressive results in the ImageNet Large Scale Visual Recognition Competition (ILSVRC) in 2012~\cite{Krizhevsky2012}. Numerous studies followed that transformed the state of the art for many computer vision applications. Even though CNNs have existed for a couple of decades already~\cite{lecun1998gradient}, this breakthrough was only made possible thanks to the current processing capabilities and the large image databases available. The potential of deep learning in medical image analysis is already being investigated and the initial results are promising~\cite{Greenspan20161153}. However, the adaptation of the existing deep learning tools from the domain of natural color images to medical images brings new challenges.

Firstly, medical imaging data are much more difficult to acquire compared to general imagery, which is freely available on the Internet. On top of that, their annotation has to be performed by multiple specialists to ensure its validity, whereas in natural image recognition anyone could serve as annotator. This lack of data makes the training on medical images very difficult or even impossible for many of the huge networks proposed in computer vision. A common way to overcome this problem is to pretrain the networks on large color image databases like ImageNet, and then fine-tune them on medical imaging data, a method often referred to as transfer learning. This approach has yielded adequately good results for many applications and has demonstrated the effectiveness of transfer learning between rather different image classification tasks~\cite{Greenspan20161153}. Secondly, the architecture of popular CNNs from the field of computer vision, is generally suboptimal for problems encountered in medical imaging such as texture analysis, while their input size is fixed and often not suitable.

To deal with these issues, in~\cite{Anthimop2016} we proposed a novel CNN that achieved significant improvement with respect to the state of the art. The network's architecture was especially designed to extract the textural characteristics of ILD patterns, while its much smaller size allowed it to be successfully trained on solely medical data without transfer learning. In this study, we propose a novel training approach that improves the performance of the newly introduced CNN, by additionally exploiting relevant knowledge, transferred from multiple general texture databases.

\section{Related Work}

\begin{table*}[!ht]
\caption{Description of the source domain databases}
\label{tab:datasets}

\begin{minipage}{\linewidth}
\begin{center}
\begin{tabular}{@{}c|cccccc|cc@{}}

\toprule

Database &
\begin{tabular}[c]{@{}c@{}}Type\end{tabular} &
\begin{tabular}[c]{@{}c@{}}Number \\ of classes\end{tabular} &
\begin{tabular}[c]{@{}c@{}}Number of\\ instances \\ per class\end{tabular} &
\begin{tabular}[c]{@{}c@{}}Number of \\ images \\ per instance\end{tabular} &
\begin{tabular}[c]{@{}c@{}}Total number\\ of images\end{tabular} &
\begin{tabular}[c]{@{}c@{}}Area per image\\ ($10^3$px)\end{tabular} &
\begin{tabular}[c]{@{}c@{}}Number of\\training patches\end{tabular} &
\begin{tabular}[c]{@{}c@{}}Number of\\validation patches\end{tabular} \\

\midrule

ALOT~\cite{burghouts2009material} & Color & 250 & 1 & \begin{tabular}[c]{@{}l@{}}100\end{tabular} & 25000 & 98.304                                                             & 257880 & 85870 \\

\midrule

DTD~\cite{cimpoi14describing} & Color & 47 & 120 & \begin{tabular}[c]{@{}l@{}}1\end{tabular} & 5640 & 229.95 $\pm$ 89.14 & 180351 & 87485 \\

\midrule

FMD~\cite{sharan2009material} & Color & 10 & 100 & \begin{tabular}[c]{@{}l@{}}1\end{tabular} & 1000 & 158.3 $\pm$ 43.2 & 18247 & 6285 \\

\midrule

KTB~\cite{Kylberg2011c} & Grey & 27 & 160 & 1 & 4480 & 331.776 & 207360 & 69120 \\

\midrule

KTH-TIPS-2b~\cite{mallikarjuna2006kth} & Color & 11 & 4 & \begin{tabular}[c]{@{}l@{}}108\end{tabular} & 4752 & 40 & 31481 & 10410 \\

\midrule

UIUC~\cite{lazebnik2005sparse} & Grey & 25 & 40 & 1 & 1000 & 307.2 & 47250 & 15750  \\



\bottomrule

\end{tabular}
\end{center}
\end{minipage}
\end{table*}

In this section, we provide a brief overview of the previous studies on ILD pattern classification, followed by a short introduction to transfer learning using CNNs.

\subsection{ILD Pattern Classification}
\label{related_work:ild}

A typical ILD pattern classification scheme takes as input a local region of interest (ROI) or volume of interest (VOI), depending on the available CT imaging modality, and is mainly characterized by the chosen feature set and classification method. The first proposed systems used handcrafted texture features such as first order statistics, gray level co-occurrence matrices, run-length matrices and fractal analysis~\cite{uppaluri1999computer}. Other systems utilized filter banks~\cite{sluimer2003computer, anthimopoulos2014classification}, morphological operations~\cite{uchiyama2003quantitative}, wavelet transformations~\cite{vo2010multiple} and local binary patterns~\cite{sorensen2010quantitative}. Moreover, the ability of multiple detector computed tomography (MDCT) scanners to provide three-dimensional data has motivated researchers to expand existing 2D texture descriptors to three dimensions~\cite{zavaletta2007high, korfiatis2010texture,depeursinge2015optimized}. More recently, researchers proposed the use of feature sets learned from data, which are able to adapt to a given problem. Most of these methods rely on unsupervised techniques, such as bag of features~\cite{gangeh2010texton, foncubierta2011using} and sparse representation models~\cite{zhao2013classification, vo2011multiscale}. Restricted Boltzmann machines (RBM) have also been used~\cite{li2013lung} to learn multi-scale filters with their responses being used as features. Once the feature vector of a ROI or VOI has been calculated, it is fed to a classifier that is trained to discriminate between the patterns. Many different approaches have been proposed for classification, including linear discriminant analysis~\cite{sluimer2003computer} and Bayesian~\cite{uppaluri1999computer} classifiers, k-nearest neighbors~\cite{sorensen2010quantitative,korfiatis2010texture}, fully-connected artificial neural networks~\cite{uchiyama2003quantitative}, random forests ~\cite{anthimopoulos2014classification} and support vector machines~\cite{depeursinge2012near,gangeh2010texton}.

Some attempts have recently been made to use deep learning techniques and especially CNNs for the classification of lung tissue patterns. Unlike the aforementioned feature learning methods, CNNs learn features in a supervised manner and train a classifier at the same time, by minimizing a cost function. Although the term deep learning refers to multiple learning layers, the first studies on the problem utilized rather shallow architectures. A modified RBM was proposed in~\cite{van2016combining} that resembles a CNN and performs both feature extraction and classification. Hidden nodes share weights and are densely connected to output nodes, while the whole network is trained in a supervised manner. In~\cite{li2014medical}, a CNN with one convolutional layer and three fully-connected layers was used, but the rather shallow architecture of the network was unable to capture complex non-linear image characteristics. In our previous work~\cite{Anthimop2016}, we designed and trained for the first time (to the best of our knowledge) a deep CNN for the problem of lung tissue classification, which outperformed shallower networks. The proposed CNN consists of 5 convolutional layers with 2 $\times$ 2 kernels and LeakyReLU activations, followed by global average pooling and three fully-connected layers. Other studies have used popular deep CNNs that exploit the discriminative power gained by pretraining on huge natural image datasets ~\cite{Shin2016}. Although the architecture of these networks is far from optimal for lung tissue classification, they managed to achieve relatively good results by transferring knowledge from other tasks.

\subsection{Transfer Learning}
\label{related_work:knowledge_transferring}

Transfer learning is generally defined as the ability of a system to utilize knowledge learned from one task, to another task that shares some common characteristics. Formal definitions and a survey on transfer learning can be found in~\cite{pan2010survey}. In this study, we focus on supervised transfer learning with CNNs. Deep CNNs have shown remarkable abilities in transferring knowledge between apparently different image classification tasks or even between imaging modalities for the same task. In most cases, this is done by weight transferring. A network is pretrained on a source task and then the weights of some of its layers are transferred to a second network that is used for another task. In some cases, the activations of this second network are just used as ``off-the-shelf'' features which can then be fed to any classifier~\cite{Razavian} . In other cases, the non-transferred weights of the network are randomly initialized and a second training phase follows, this time on the target task~\cite{Yosinski2014}. During this training, the transferred weights could be kept frozen at their initial values or trained together with the random weights, a process usually called ``fine-tuning''. When the target dataset is too small with respect to the capacity of the network, fine-tuning may result in overfitting, so the features are often left frozen. Finding which and how many layers to transfer depends on the proximity of the two tasks but also on the proximity of the corresponding imaging modalities. It has been shown that the last layers of the network are task specific while the earlier layers of the network are modality specific~\cite{Castrejon_2016_CVPR}. On the other hand, if there are no overfitting issues, the best strategy is to transfer and fine-tune every layer~\cite{Yosinski2014}. This way, the discovered features are adapted on the target task, while keeping the useful common knowledge. Another type of transfer learning is the multi-task learning (MTL) approach that trains on multiple related tasks simultaneously, using a shared representation~\cite{Caruana1997MTL}. Such process may increase the performance for all these tasks and It is typically applied when training data for some tasks are limited.

Transfer learning has been extensively studied over the past few years, especially in the field of computer vision, with several interesting findings. In~\cite{zheng2016good}, pretrained CNNs such as VGG-Net and AlexNet are used to extract ``off-the-shelf'' CNN features for image search and classification. The authors demonstrate that fusing features extracted from multiple CNN layers improves the performance on different benchmark databases. In~\cite{azizpour2015feature}, the factors that influence the transferability of knowledge in a fine-tuning framework are investigated. These factors include the network's architecture, the resemblance between source and target tasks and the training framework. In a similar study~\cite{Yosinski2014}, the effects of different fine-tuning procedures on the transferability of knowledge are investigated, while a procedure is proposed to quantify the generality or specificity of a particular layer. A number of studies have also utilized transfer learning techniques, in order to adapt well-known networks to classify medical images. In most of the cases, the network used is the VGG, AlexNet or GoogleNet pretrained on ImageNet~\cite{Shin2016},~\cite{tajbakhsh2016convolutional}. However, these networks are designed with a fixed input size usually of 224 $\times$ 224 $\times$ 3, so that  images have to be resized and their channels artificially extended to three, before being fed to the network. This procedure is inefficient and may also impair the descriptive ability of the network.

\section{Materials \& Methods}
\label{sec:methods}
\begin{figure*}
    \centering
    \begin{subfigure}[b]{1\textwidth}
    	\centering
        \includegraphics[width=0.9\textwidth,natwidth=610,natheight=642]{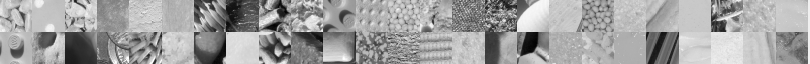}
    \end{subfigure}
    ~

    \begin{subfigure}[b]{1\textwidth}
    	\centering
        \includegraphics[width=0.9\textwidth,natwidth=610,natheight=642]{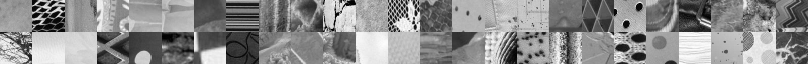}
    \end{subfigure}
    ~

    \begin{subfigure}[b]{1\textwidth}
    	\centering
        \includegraphics[width=0.9\textwidth,natwidth=610,natheight=642]{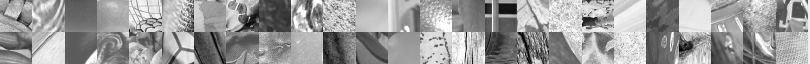}
    \end{subfigure}
    ~

    \begin{subfigure}[b]{1\textwidth}
    	\centering
        \includegraphics[width=0.9\textwidth,natwidth=610,natheight=642]{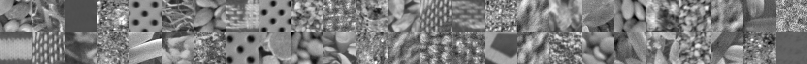}
    \end{subfigure}
    ~

    \begin{subfigure}[b]{1\textwidth}
    	\centering
        \includegraphics[width=0.9\textwidth,natwidth=610,natheight=642]{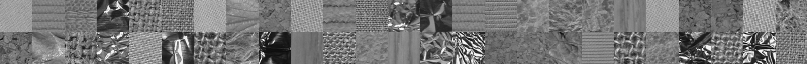}
    \end{subfigure}
    ~

    \begin{subfigure}[b]{1\textwidth}
    	\centering
        \includegraphics[width=0.9\textwidth,natwidth=610,natheight=642]{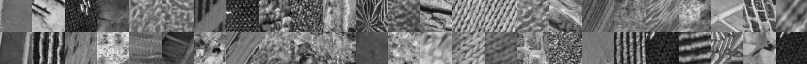}
    \end{subfigure}
    ~

    \begin{subfigure}[b]{1\textwidth}
    	\centering
        \includegraphics[width=0.9\textwidth,natwidth=610,natheight=642]{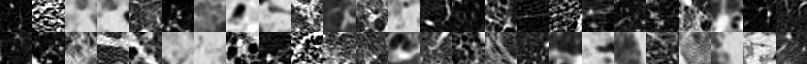}
    \end{subfigure}

    \caption{Typical samples from each dataset. The color databases were converted to gray scale. From top to bottom: ALOT, DTD, FMD, KTB, KTH-TIPS-2b, UIUC, ILD}
    \label{fig:samples}
\end{figure*}

In this section we present a method for transferring knowledge from multiple source databases to a CNN, ultimately used for ILD pattern classification. Prior to this, we describe the databases that were utilized for the purposes of this study as well as the architecture of the newly proposed CNN, in order to provide a better foundation for the description of the methodology.

\subsection{Databases}
\label{sec:databases}

Six texture benchmark databases were employed to serve as source domains for the multi-source transfer learning: the Amsterdam library of Textures (ALOT)~\cite{burghouts2009material}, the Describable Textures Dataset (DTD)~\cite{cimpoi14describing}, the Flickr Material Database (FMD)~\cite{sharan2009material}, Kylberg Texture Database (KTB)~\cite{Kylberg2011c}, KTH-TIPS-2b~\cite{mallikarjuna2006kth} and the Ponce Research Group's Texture database (UIUC)~\cite{lazebnik2005sparse}. Moreover, the concatenation of all aforementioned databases was also used. As target domain, we used two databases of ILD CT scans from two Swiss university hospitals: the Multimedia database of ILD by the University Hospital of Geneva (HUG)~\cite{depeursinge2012building} and the Bern University Hospital, ``Inselspital'' (Insel) database~\cite{Anthimop2016}.

\subsubsection{Source Domain Datasets}
All the source domain databases are publicly available texture classification benchmarks. Each class corresponds to a specific texture (e.g. fabric, wood, metal, foliage) and is represented by pictures of one or more instances of the texture. Two of the databases -- ALOT and KTH-TIPS-2b -- also contain multiple pictures for each instance under different angles, illumination and scales. The image size is fixed for all databases apart from DTD, while FMD also provides texture masks. 

For the creation of the training-validation dataset, all the color databases (i.e. ALOT, DTD, FMD, KTH-TIPS-2b) were converted to gray-scale and non-overlapping patches were extracted with a size equal to the input of the proposed CNN namely, $32 \times 32$. When not provided, partitioning between training and validation sets was performed at the instance level, except for ALOT, where the number of instance is equal to the number of classes. No testing set was created for the source domain databases, since the ultimate goal is to test the system only on the target domain. In the case of DTD, where training, validation and test sets are provided, the test set was added to the training set. Table~\ref{tab:datasets} summarizes the characteristics of the original source databases and the corresponding patch datasets.

\subsubsection{Target Domain Dataset}

The HUG database~\cite{depeursinge2012building} consists of 109 HRCT scans of different ILD cases with 512 $\times$ 512 pixels per slice and an average of 25 slices per case. The average pixel spacing is 0.68mm, and the slice thickness is 1-2mm. Manual annotations for 17 different lung patterns are also provided, along with clinical parameters from patients with histologically proven diagnoses of ILDs. The Insel database consists of 26 HRCT scans of ILD cases with resolution 512 $\times$ 512 and an average of 30 slices per case. Average pixel spacing is 0.62mm and slice thickness is 1-2mm.

A number of preprocessing steps was applied to the CT scans before creating the final ILD patch dataset. The axial slices were rescaled to match a certain x,y-spacing value that was set to 0.4mm, while no rescaling was applied on the z-axis. The image intensity values were cropped within the window [-1000, 200] in Hounsfield units (HU) and mapped to [0, 1]. Experienced radiologists from Bern University hospital annotated (or re-annotated) both databases by manually drawing polygons around seven different patterns including healthy tissue and the six most relevant ILD patterns, namely ground glass, reticulation, consolidation, micronodules, honeycombing and a combination of ground glass and reticulation. In total 5529 ground truth polygons were annotated, out of which 14696 non-overlapping image patches of size 32 $\times$ 32 were extracted, unequally distributed across the 7 classes. The patches are entirely included in the lung field and they have an overlap with the corresponding ground truth polygons of at least 80\%. From this patch dataset, 150 patches were randomly chosen from each class for the validation and 150 for the test set. The remaining patches were used as the training set, which was artificially augmented to increase the amount of training data and prevent over-fitting. Label-preserving transformations were applied, such as flip and rotation, as well as combinations of the two. In total, 7 transformations were used while duplicates were also added for the classes with few samples. The final number of training samples was constrained by the rarest class and the condition of equal class representation that led to 5008 training patches for each class. In total, the training set consists of 35056 patches while the validation and test sets contain of 1050 patches each. More details about this dataset can be found in~\cite{Anthimop2016}.

\subsection{CNN Architecture}
\label{sec:cnn}

In order to minimize the parameters involved and focus only on the aspects of transfer learning, we used the same CNN architecture as proposed in~\cite{Anthimop2016} throughout the different steps of the method. The input of the network is an image patch of 32 $\times$ 32 pixels. This patch is convolved by five subsequent convolutional layers with 2 $\times$ 2 kernels, while the number of kernels is proportional to the receptive field of each layer with respect to the input. The number of kernels we used for the $L_{th}$ layer is $k(L+1)^{2}$, where the parameter $k$ depends on the complexity of the input data and was chosen to be 4. The output of the final convolutional layer is globally pooled, thus passing the average value of each feature map to a series of three dense layers. A rectified linear unit (ReLU) is used as the activation function for the dense layers, while the convolutional layers employ very leaky ReLU activations with $\alpha = 0.3$. Finally, Dropout is used before each dense layer dropping 50\% of its units. For training the network, the Adam optimizer~\cite{adam2014} was used with the default values for its hyper-parameters. The training ends when none of 200 consecutive epochs improves the network's performance on the validation set by at least 0.5\%. 

\subsection{Multi-source Transfer learning}
\label{sec:proposed_method}

\begin{figure}
	\centering
	\includegraphics[width=0.4\textwidth]{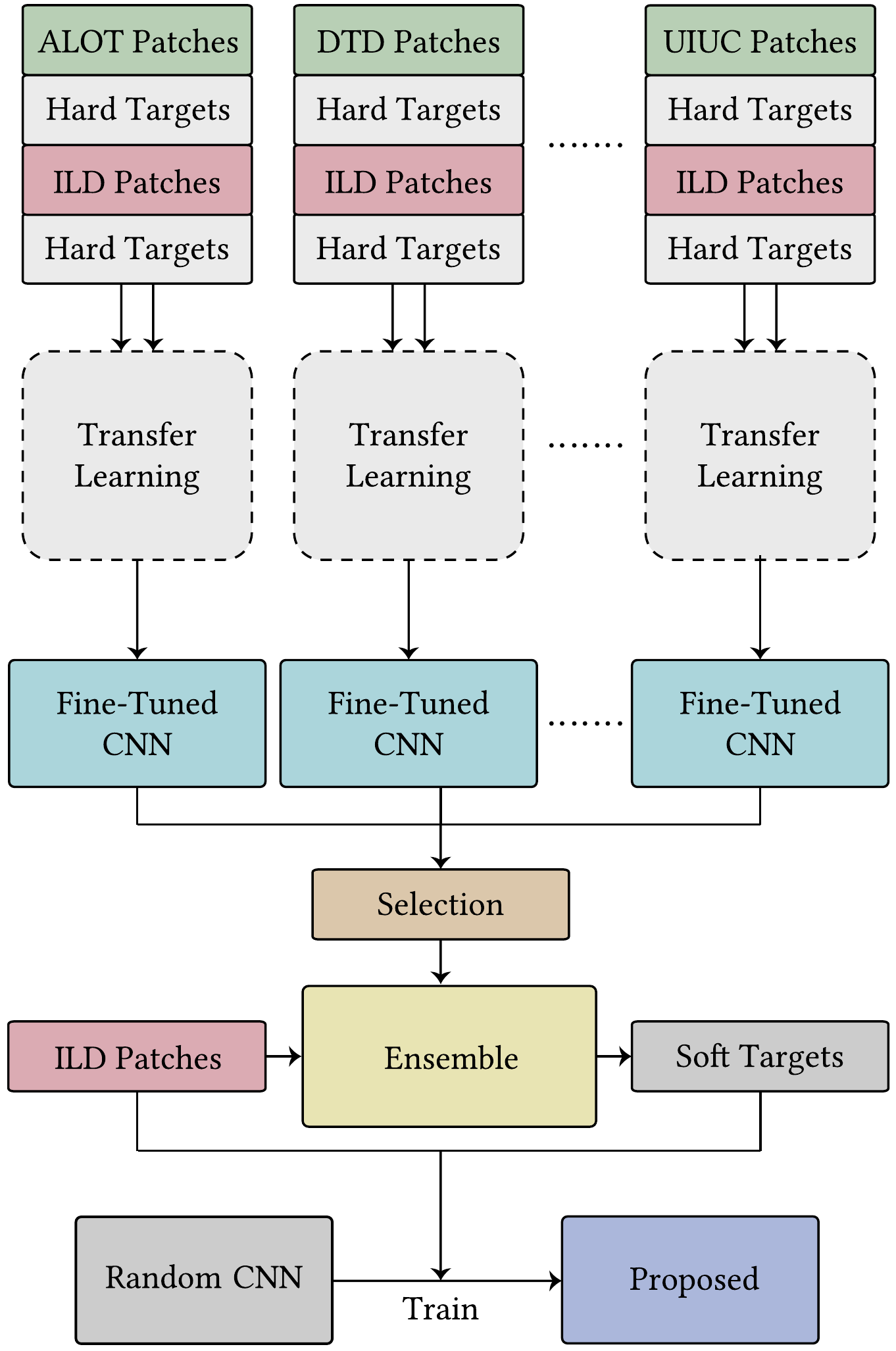}
	\caption{Multi-source Transfer Learning:  Knowledge is transferred from each source database to a different CNN. A selection process combines CNNs into an ensemble that is used to teach a single randomly initialized model.}
	\label{fig:rTL}
\end{figure}

The source datasets presented in Section~\ref{sec:databases} demonstrate a wide spectrum of different characteristics, as shown in Fig.~\ref{fig:samples} and Table~\ref{tab:datasets}; hence, we expect that they will also contribute a range of diverse and complementary features. If this assumption holds, the parallel transfer learning from all datasets into one model will improve its performance more than any individual dataset would. However, the standard transfer learning approach by transferring weights can only utilize one source dataset. To tackle this problem, we transfer knowledge from each source to a different CNN and then fuse them into an ensemble that is expected to have performance superior to any of the individual models but also a larger computational complexity. We then transfer the fused knowledge back to a network with the original architecture, in order to reduce the complexity while keeping the desirable performance. Simple weight transferring is again not possible here, since it requires models with the same architecture. We therefore use model compression, a technique that transfers knowledge between arbitrary models for the same task. Fig.~\ref{fig:rTL} depicts the full pipeline of the proposed multi-source transfer learning method while in the next paragraphs, we describe its three basic components in more detail.

\subsubsection{Single-Source Transfer Learning}
\label{sec:transfer_learning}
\begin{figure}
	\centering
	\includegraphics[width=0.25\textwidth]{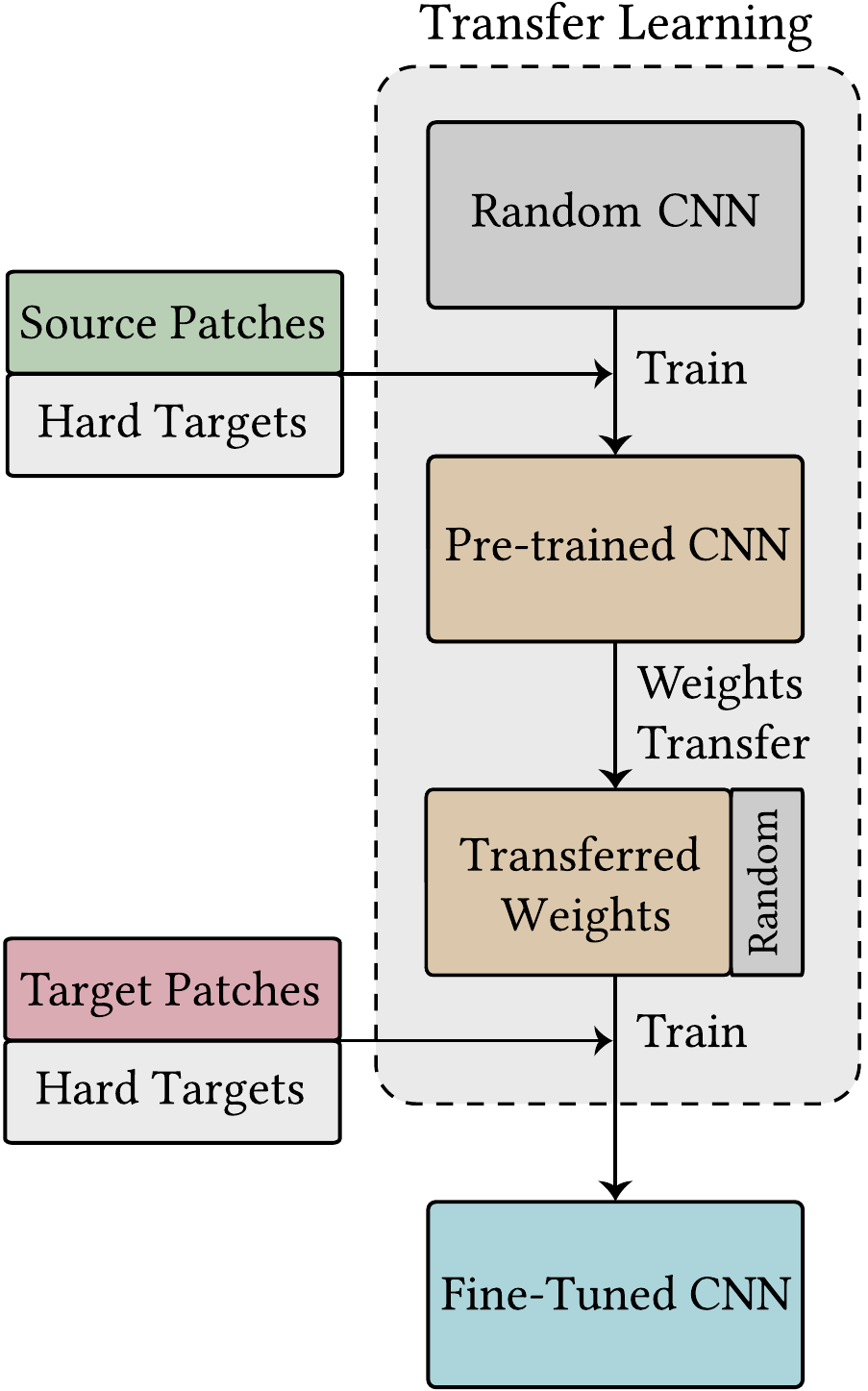}
	\caption{Transfer Learning through weight transfer}
	\label{fig:sTL}
\end{figure}

Fig.~\ref{fig:sTL} illustrates the used weight transfer scheme from a source task to the target task, namely the ILD classification. Starting from the first layer, a number of consecutive layers are transferred from the pretrained network to initialize its counterpart network. The rest of the network is randomly initialized, while the last layer changes size to match the number of classes in the target dataset (i.e. 7). The transferred layers are then fine-tuned along with the training of the randomly initialized ones. We decided to fine-tune the layers instead of freezing them since the proposed network is relatively small and has been previously trained on the target dataset without overfitting~\cite{Anthimop2016}. According to~\cite{Yosinski2014} weight freezing should only be used to avoid overfitting problems. In order to investigate the effects of transferring different number of layers, we have performed a set of experiments for each of the source datasets.

\subsubsection{Knowledge Fusion in an Ensemble}
\label{sec:ensemble}
Ensembles are systems that use multiple predictors, statistically independent to some extent, in order to attain an aggregated prediction. Using ensembles to achieve a better performance is a well-established technique and has been successfully exploited in many applications~\cite{Dietterich2000}. Such systems usually perform better than each of the predictors alone, while they also gain stability. This performance gain arises from the fact that the different prediction models that form the ensemble, capture different characteristics of the function to be approximated.

In order to build a strong ensemble, instead of manually selecting the models, we implemented an ensemble selection approach similar to the one presented in~\cite{caruana2004ensemble}. The employed algorithm is a forward selection procedure which selects models from a pool and iteratively adds them to the ensemble following a specific criterion. Moreover, some additions to prevent over-fitting were also implemented. The pool from which the algorithm selects models includes all the networks that were pretrained on the source datasets and fine-tuned on the ILD dataset, snapshots of these networks during training, as well as a few randomly initialized networks trained from scratch on the target data. After creating the CNN model pool, a subset is randomly sampled from it with half of its size. Then, the models in the subset are ranked by their performance and the best $N$ models are chosen to initialize the ensemble. From the rest of the subset's models, we add the one that increases the ensemble performance the most, and continue adding models until no more gain can be achieved. Model selection is performed with replacement, meaning that the same model can be included more than once. The whole procedure is repeated for $M$ subsets generating $M$ ensembles which are then aggregated into one, by averaging their outputs. The selection of the models is based on the average F-score of the validation set while the involved parameters have been tested on a grid with $N=\{1,2,\ldots,25\}$ and $M=\{1,2,\ldots,15\}$. For each position of the parameter grid the selection was repeated 100 times and finally the best ensemble was found for $N=2$ and $M=5$.

\subsubsection{Model Compression}

Model compression is used as a final step, to compress the knowledge of the huge ensemble created by the previous procedure, into a single network with the original architecture. Model compression, also known as knowledge distillation, is the procedure for training a model using ``soft targets'' that have been produced by another, usually more complex model~\cite{BuciluaModCompression}~\cite{Hinton2015distillation}. As soft targets one can use the class probabilities produced by the complex model or the logits namely, the output of the model before the softmax activation. The model that produces the soft targets is often called the teacher, while the model to which the knowledge is distilled plays the role of the student. The soft targets carry additional knowledge discovered by the teacher, regarding the resemblance of every sample to all classes. This procedure can be considered as another type of knowledge transfer which is performed for the same task, yet between different models. In our case, the ensemble is employed as a teacher while the student is a single, randomly initialized CNN with the original architecture described in Section~\ref{sec:cnn}. After being trained on the soft targets the student model will approximate the behavior of the ensemble model and will even learn to make similar mistakes. However, these are mistakes that the student would have probably made by training on the hard targets, considering its relatively inferior capacity.

\subsection{Multi-task Learning}
\label{sec:MTL}
\begin{figure}
	\centering
	\includegraphics[width=0.35\textwidth]{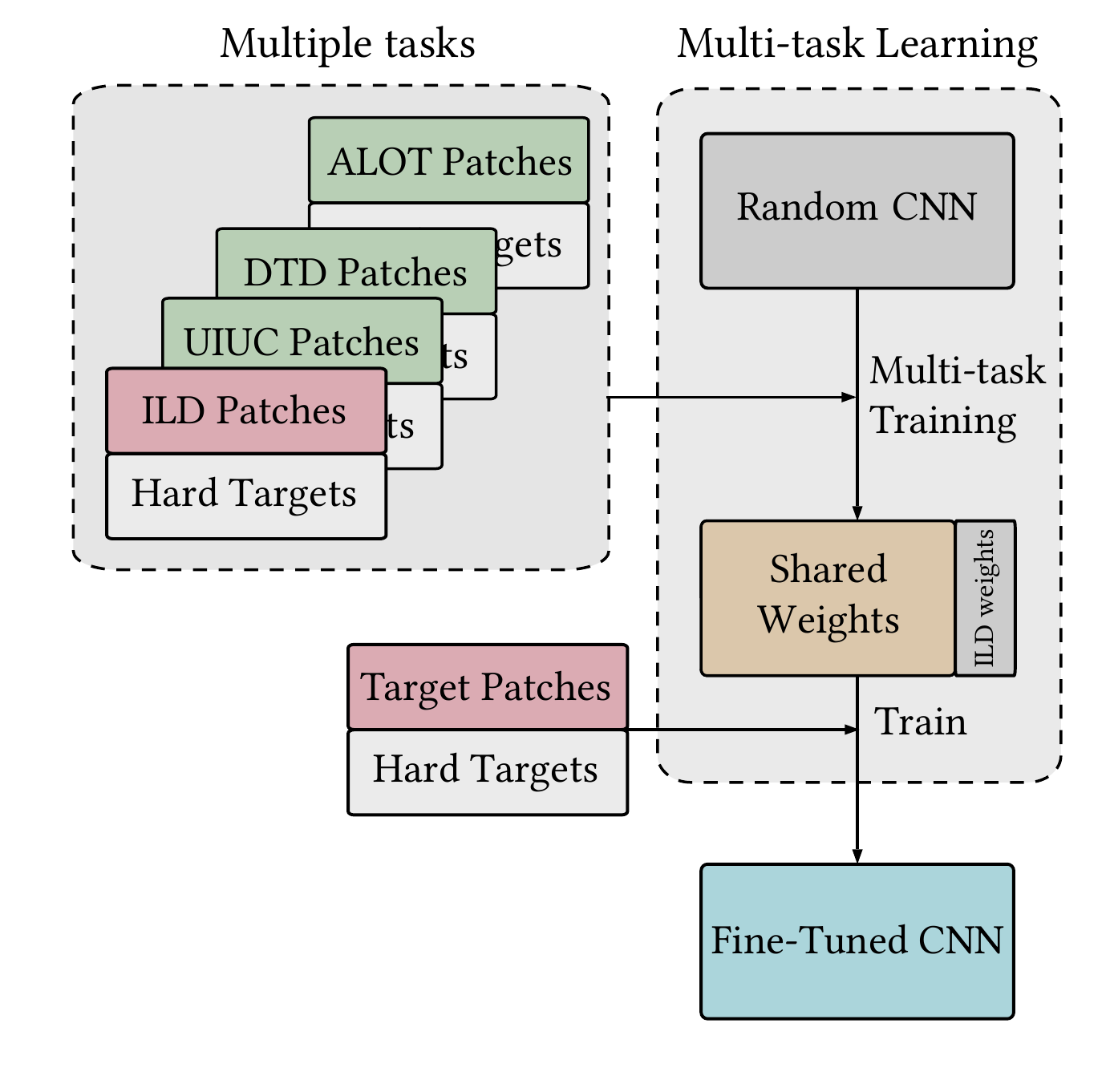}
	\caption{Multi-task Learning}
	\label{fig:mTL}
\end{figure}

MTL is another way to fuse knowledge from multiple sources into multiple models. In this study we used it as a baseline method. The method simultaneously trains models for each of the tasks, with some of the weights shared among all models. In our implementation, we train seven networks, one for each of the source datasets and one for the target dataset. These CNNs share all the weights apart from the last layer, the size of which depends on the number of classes for that particular task. The parallel training was achieved by alternating every epoch the task between the target and one of the source tasks. In other words, odd epochs train on the target task while even epochs train on source tasks in a sequential manner. Although MTL fuses knowledge from all involved tasks, it does not use tasks exclusively as source or target like the standard transfer learning approach. Since our final goal is to improve the performance of the target task, we further fine-tune the resulting model on the ILD dataset. Fig.~\ref{fig:mTL} depicts an outline of our multi-task learning approach.

\section{Experimental Setup \& Results}
\label{sec:results}
In this section we describe the setup of the conducted experiments, followed by the corresponding results with the related discussion.

\subsection{Experimental Setup}
For all the experiments presented in this section, a train-validation-test scheme was utilized. The presented results were calculated on the test set while the selection of hyper-parameters and the best resulting models was made over the validation set. In the rest of this section, we describe the chosen evaluation protocol and some implementation details.

\subsubsection{Evaluation}
As a principle evaluation metric we used the average F$_1$-score over the different classes, due to its increased sensitivity to imbalances among the classes. The F$_{1}$-score is calculated as follows:

$$
F_{avg} = \frac{2}{7} \sum_{c=1}^{7} \frac{recall_{c} \cdot precision_{c}}{recall_{c}+precision_{c}}
$$

where $recall_{c}$ is the fraction of samples correctly classified as $c$ over the total number of samples of class $c$, and the $precision_{c}$ is the fraction of samples correctly classified as $c$ over all the samples classified as $c$.

\subsubsection{Implementation}
The proposed method was implemented in Python using the Keras~\cite{chollet2015keras} framework with a Theano~\cite{theano2016arXiv} back-end. All experiments were performed under Linux OS on a machine with CPU Intel Core i7-5960X @ 3.50GHz, GPU NVIDIA GeForce Titan X, and 128GB of RAM.

\subsection{Results}
In this section, we present the results of the performed experiments, grouped according to the three basic components of the system as presented in Section~\ref{sec:proposed_method}. Finally, we analyze the performance of the proposed network and compare with other methods.

\subsubsection{Single-Source Transfer Learning}
\label{sec:singe-source-transfer}
In this first series of experiments we investigate the performance gain by transferring knowledge from individual source datasets to the target task, i.e. the classification of ILD patterns. A CNN model was pretrained on each of the six source datasets and then fine-tuned on the ILD data. A seventh source dataset was added that consists of all six datasets merged in one. As described in Section~\ref{sec:cnn}, the proposed network has five convolutional and three dense layers. Starting from the first, we transfer one to seven layers for each of the pretrained networks. The rest of the layers are randomly initialized and the entire CNN is fine-tuned on the ILD task. Different random initializations may result in deviations of the results so to minimize this effect, we repeated each experiment three times and reported the mean values. 

\begin{figure}
	\centering
	\includegraphics[width=0.5\textwidth]{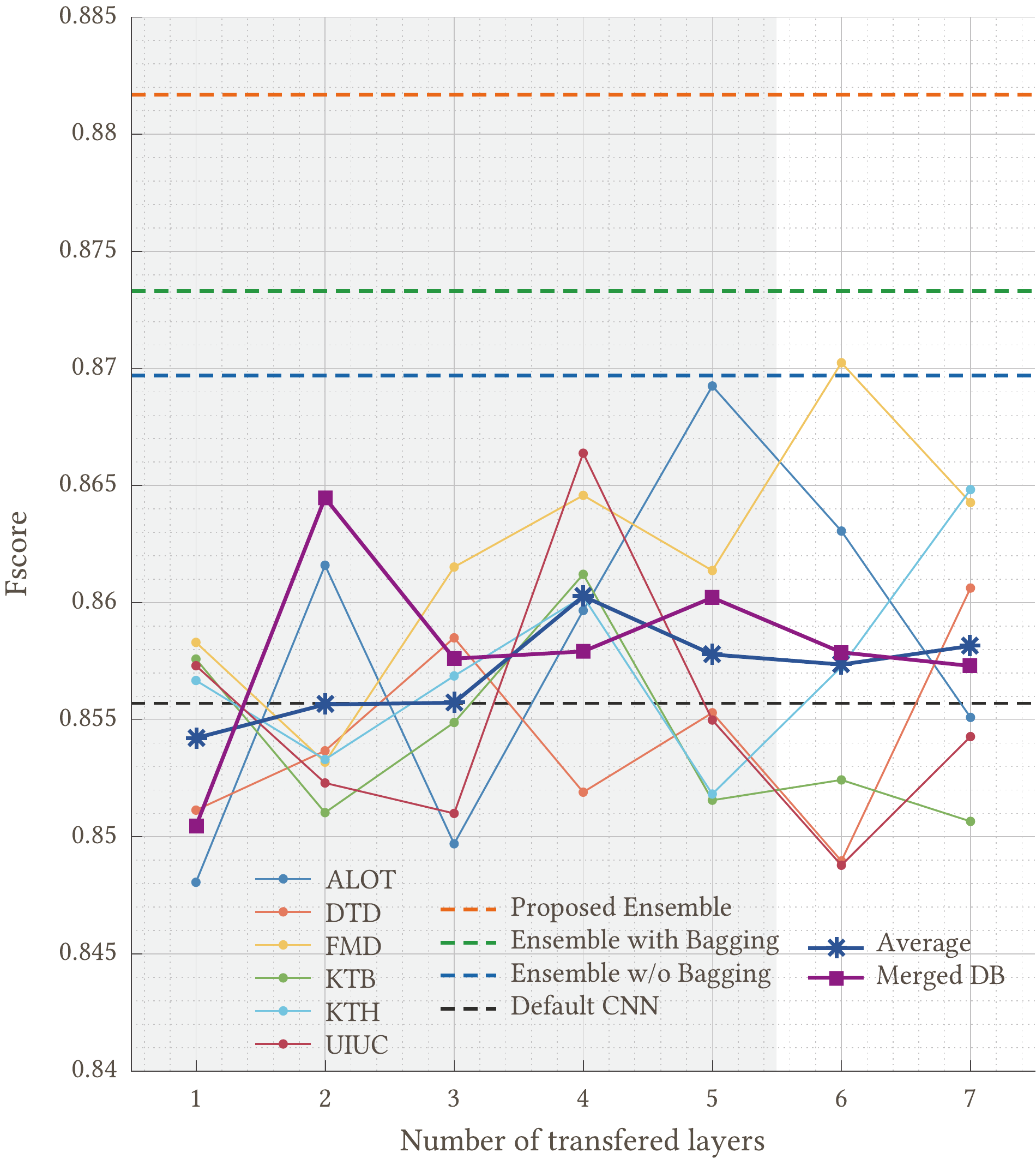}
	\caption{The F$_{1}$-score produced by transferring of knowledge from single source domains for different number of transferred layers, averaged over three experiments. The horizontal lines correspond to the CNN without knowledge transfer or the different ensembles of CNNs.}
	\label{fig:pre-fine}
\end{figure}

The results of this experiment are depicted in Fig.~\ref{fig:pre-fine}, where the region of the light gray background denotes the convolutional layers, while the rest denote the first two dense layers. The horizontal dashed line at 0.855 represents the performance of the network trained from scratch (with random initialization). The best results were achieved when six layers (i.e. five convolutional layers and one dense) were transferred from the CNN that was pretrained on the FMD dataset. However, no optimal weight transferring strategy can be inferred for every pretrained network, due to their relative different behavior. An additional line with the average performance over all source datasets is also shown. According to this line, the contribution of weight transferring increases, on average, when transferring at least four layers. Weight transferring seems to help even when transferring all layers. This is probably due to the ability of fine-tuning to adapt even the most task-specific features to the target task, an observation which is inline with the conclusions of~\cite{Yosinski2014}. 

As for the runtime of the experiments, one could expect a faster training for a pretrained network since its initial state is closer to a good solution than a randomly initialized network. Indeed, the average number of epochs for the pretrained is 426 instead of 479 for the random, with each epoch taking about 12 seconds. However, this difference is small and statistically non-significant ($p \approx 0.11$) probably due to the fact that loss drops with a lower rate while approaching the end of training, so the starting point does not significantly affect the number of required epochs.

The conducted experiments have demonstrated that the random initializations before pretraining or fine-tuning, as well as the different source datasets may introduce a significant variance between the network's results. This unstable behavior of single-source transfer learning combined with the assumption of reduced correlation among the resulting models, motivated us to build an ensemble model to fuse the extracted knowledge and reduce the aforementioned variance.

\subsubsection{Knowledge Fusion in an Ensemble}
Fig.~\ref{fig:pre-fine} also illustrates the performance of the ensemble that was built as described in Section~\ref{sec:ensemble}. The ensemble clearly outperforms the rest of the models by reducing their variance (through output averaging) and by transferring multi-source knowledge, at the same time. In order to investigate the contribution of ensemble averaging alone, we also built an ensemble from a pool of randomly initialized models. The output of this ensemble reached a performance of 0.8697 which is better than the single randomly initialized CNN but still inferior to the multi-source ensemble. In addition, we used bootstrap aggregating (bagging) to boost the performance even more by reducing the correlation between the models. To this end, we trained each CNN of the ensemble on samples randomly sampled from the training set with replacement. The performance was slightly improved reaching 0.8733 which was however still inferior to the proposed ensemble. These results showed that although the ensemble by itself may increase the accuracy of stochastic models, the transferred knowledge also contributes to the final result.

\subsubsection{Model Compression}
For this last part, the ensemble was employed as a teacher producing soft targets for the ILD training dataset that were then used to train CNNs. We experimented with a number of different choices for the student networks choosing between the pretrained and fine-tuned networks from the previous steps as well as randomly initialized ones. All of the different students reached similar levels of performance, so we finally chose as student the one with the random initialization, for simplicity. The achieved performance after teaching the chosen student was 0.87518 in the test set. This result lies below the ensemble's performance yet above all the previously presented results.

\subsubsection{Performance and Comparison with Previous Work}
As a baseline method for comparison in multi-source transfer learning we used an MTL approach as described in Section~\ref{sec:MTL}. The performance on the ILD task while training along with the other tasks only reached the value of 0.8110. After a fine-tuning step, the performance reached the value of 0.8631, which is not much better than the network trained from scratch and similar to a number of single source pretrained networks. These results could be due to the limited capacity of the network that attempts to solve multiple problems at the same time. Modifications in the MTL scheme such as weighting the contributions of the different tasks or sharing different parts of the network could yield better results, however this would increase the complexity of the scheme and would require a large number of experiments on different strategies.

\begin{figure}
	\centering
	\includegraphics[width=0.8\linewidth]{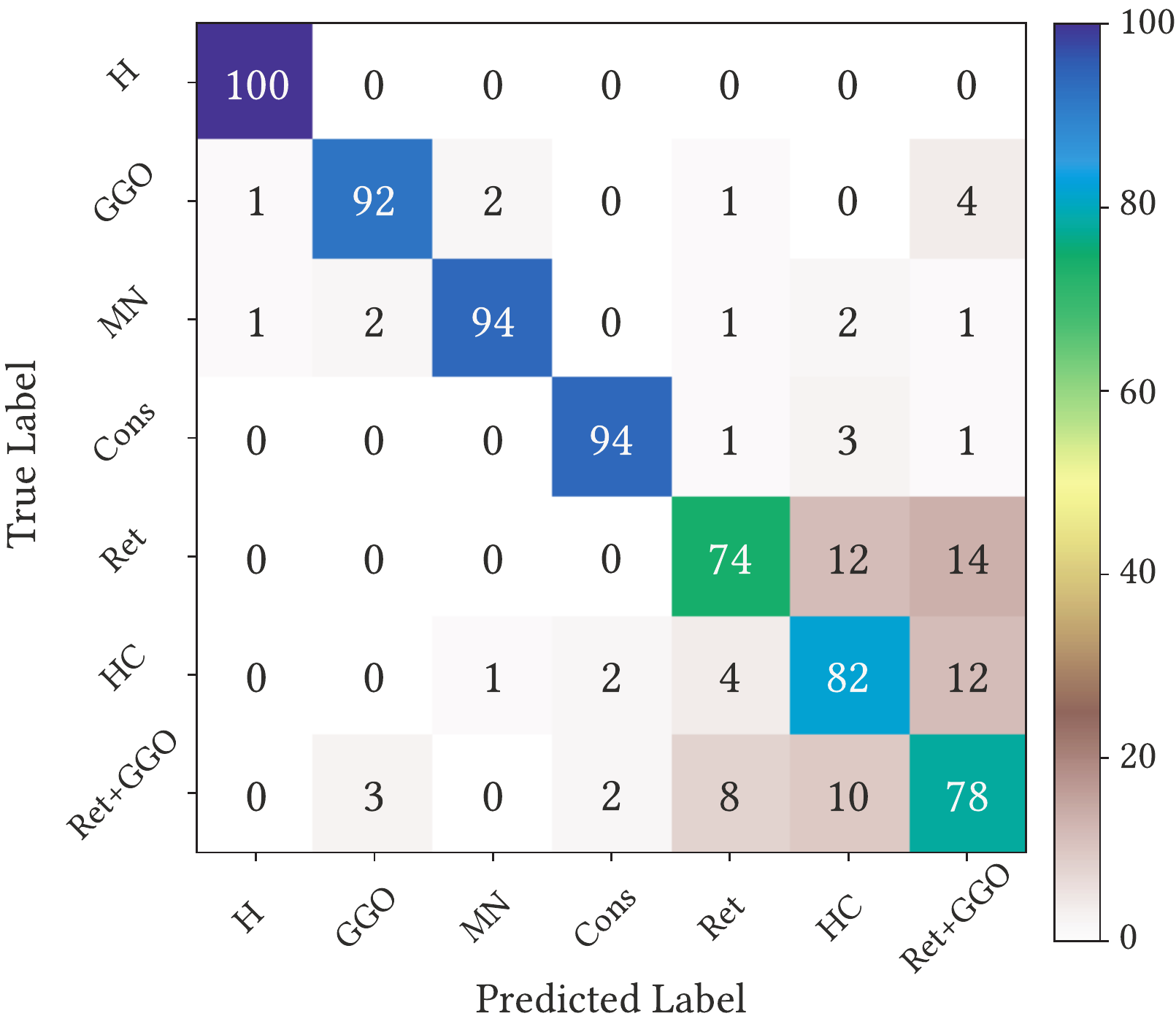}
	\caption{Confusion matrix of the proposed compressed model.}
	\label{fig:cm}
\end{figure}

Table~\ref{tab:sota} provides a comparison with other methods from the literature. The first three rows correspond to methods that use hand crafted features and a range of different classifiers. The rest correspond to methods that utilize CNNs. All the results were reproduced by the authors by implementing the different methods and using the same data and framework to test them. The proposed multi-source transfer learning technique improved the performance of the proposed network by an absolute 2\% compared to the previous performance 0.8557 of the same CNN in~\cite{Anthimop2016}. Finally, Fig.~\ref{fig:cm} shows the confusion matrix of the proposed approach. As shown, the confusion is basically between the fibrotic classes (i.e. reticulation, honeycombing and the combination of ground glass and reticulation) which was expected. One may also notice that the matrix is more balanced than the one presented in~\cite{Anthimop2016}.

\begin{table}

\centering
\caption{Comparison of the proposed method with methods from the literature}
\label{tab:sota}

\begin{tabular}{@{}ccc@{}}

\toprule
Study & Method & $F_{avg}$\\
\midrule
Gangeh \cite{gangeh2010texton}        & Local pixel textons - SVM-RBF & 0.6942 \\
Sorensen \cite{sorensen2010quantitative}     & LBP, histogram - kNN & 0.7420 \\
Anthimopoulos\cite{anthimopoulos2014classification} & \begin{tabular}[c]{@{}c@{}}Quantiles of local\\ DCT, histogram - RF\end{tabular} & 0.8170 \\
\midrule
Li \cite{li2014medical}            &  5-layer CNN  & 0.6657  \\
LeNet \cite{lecun1998gradient}      & 7-layer CNN  & 0.6783 \\
AlexNet \cite{Krizhevsky2012}  & 8-layer CNN & 0.7031 \\
Pre-trained AlexNet           & 8-layer CNN & 0.7582 \\
VGG-Net \cite{simonyan2014very}   & 16-layer CNN & 0.7804 \\
Anthimopoulos~\cite{Anthimop2016}   & 8-layer CNN & 0.8557 \\
\midrule

\bfseries Proposed Methods &  \bfseries \begin{tabular}[c]{@{}c@{}}Multi-task Learning\\ Compressed 8-layer CNN \\Ensemble of CNNs\end{tabular} & \bfseries \begin{tabular}[c]{@{}l@{}}0.8631\\ 0.8751\\0.8817\end{tabular} \\
\bottomrule

\end{tabular}

\end{table}

\section{Conclusion}

In this paper we presented a training method that improves the accuracy and stability of a CNN on the task of lung tissue pattern classification. The performance gain was achieved by the multiple transfer of knowledge from six general texture databases. To this end, a network was pretrained on each of the source databases and then fine-tuned on the target database after transferring different numbers of layers. The networks obtained were combined in an ensemble using a model selection process, which was then employed to teach a network with the original size. The resulting CNN achieved a gain in performance of about 2\% compared to the same network when trained on the hard targets. This result proves the potential of transfer learning from natural to medical images that could be beneficial for many applications with limited available medical data and/or annotations. We believe that more challenging datasets, with additional classes and/or higher diversity, may benefit even more from similar approaches. Considering that even experienced radiologists would not achieve a perfect classification, especially on a patch level, the reported performances could have reached a peak. Finally, the reported increase in accuracy comes at the expense of increased training time since multiple models have to be trained. However, the inference time is still exactly the same and the additional training time required can be considered as a fair compromise for improving the performance, in cases of data shortage. Our future research plans in the topic include the use of the ensemble teacher for labeling unlabeled samples that will augment the training set of the student model. Such an approach could partially assist with the common problem of limited annotated data in the field of medical image analysis.

\end{document}